\documentclass[runningheads]{llncs}
\usepackage{amsmath}
\usepackage{graphicx}
\usepackage{booktabs}
\usepackage{makecell}
\usepackage{wrapfig}
\usepackage{underscore}
\usepackage{hyperref}
\usepackage{paralist}
\usepackage{tikz}
\usepackage{comment}
\usetikzlibrary{positioning}
\usetikzlibrary{shapes.geometric,arrows.meta,decorations.markings}
\usetikzlibrary{calc,shapes.callouts}

\usepackage{xcolor}
\usepackage{lstsemantic}
\usepackage{lstlangmizar}

\usepackage{relsize}

\makeatletter
\newcommand\scriptsizemy{\@setfontsize\scriptsize\@viiipt\@viiipt}
\makeatother

\makeatletter
\renewcommand\section{\@startsection{section}{1}{\z@}%
                       {-12\p@ \@plus -4\p@ \@minus -4\p@}%
                       {8\p@ \@plus 4\p@ \@minus 4\p@}%
                       {\normalfont\large\bfseries\boldmath
                        \rightskip=\z@ \@plus 8em\pretolerance=10000 }}
\makeatother

\begin{document}
\title{
  First Neural
  Conjecturing\\ Datasets and Experiments\thanks{
        Funded by the \textit{AI4REASON} ERC Consolidator grant nr. 649043
        and by the Czech project AI\&Reasoning CZ.02.1.01/0.0/0.0/15\_003/0000466
        and the European Regional Development Fund. We thank K. Chvalovsk\'y and T. Gauthier for discussions.
        }}

\author{Josef Urban\inst{1} \and Jan Jakub\r{u}v\inst{1}} %

\authorrunning{Urban, Jakub\r{u}v} %

\institute{Czech Institute of Informatics, Robotics and Cybernetics, Prague,
Czech Republic }

\maketitle
\begin{abstract}
  We describe several datasets and first experiments with creating
  conjectures by neural methods.  The datasets are based on the Mizar
  Mathematical Library processed in several forms and the problems
  extracted from it by the MPTP system and proved by the E prover
  using the ENIGMA guidance. The conjecturing experiments use the
  Transformer architecture and in particular its GPT-2 implementation.

\end{abstract}

\section{Introduction and Related Work}
\label{sec:intro}
Automated creation of suitable conjectures is one of the hard problems
in automated reasoning over large mathematical corpora. This includes
tasks such as (i) conjecturing suitable intermediate lemmas (cuts) when
proving a harder conjecture, and (ii) unrestricted creation of interesting
conjectures based on the previous theory (i.e., theory exploration).
Starting with Lenat's AM~\cite{Lenat1976}, several systems
such as the more specialized Graffitti by
Fajtlowicz~\cite{Fajtlowicz88}, and Colton's HR~\cite{Colton2012}
have been developed,
typically using heuristics for theory exploration or limited
brute-force enumeration, controlled e.g. by the type system~\cite{JohanssonRSC14-short}.

Our motivation is the work  of Karpathy\footnote{\url{http://karpathy.github.io/2015/05/21/rnn-effectiveness/}} with
recurrent neural networks (RNNs). One of his experiments used the Stacks
project, generating LaTeX-style pseudo-mathematics that looked quite
credible to non-experts. We have repeated these experiments over the Mizar library using
Karpathy's RNNs in 2016, but the results did not seem
convincing. The neural methods have however improved since, coming up
with stronger methods and systems such as attention, transformer and
GPT-2~\cite{radford2019language}. The experiments described here started by testing GPT-2 on
the Mizar library, gradually producing several more datasets.

Related work includes research on the informal-to-formal
grammar-based and neural translation~\cite{KaliszykUV15-short,DBLP:conf/itp/KaliszykUV17-short,WangKU18-short,WangBKU20-short}. There
it was found that PCFGs and RNNs with attention work well on
some informal-to-formal datasets, can learn analogies from the data,
and can be used to produce multiple formal outputs of which some 
are new provable conjectures. In~\cite{WangBKU20-short} we use this
together with type checking to set up a data-augmentation loop between
the neural learner and the type-checker. Such learning-reasoning loops are also planned for the datasets presented here.
Similar %
experiments are done in
~\cite{GauthierKU16-short} and by
Chvalovsk\'y\footnote{\url{http://aitp-conference.org/2019/abstract/AITP_2019_paper_27.pdf}, \url{http://aitp-conference.org/2020/abstract/paper_21.pdf}}.
Gauthier has been working on term synthesis using Monte-Carlo
Tree Search and reinforcement learning with semantic feedback~\cite{DBLP:journals/corr/abs-1912-01525,DBLP:journals/corr/abs-1910-11797}.

\section{Datasets}

The datasets for neural conjecturing are available from our web page\footnote{\url{http://grid01.ciirc.cvut.cz/~mptp/nn_conj20/}}.
We have so far experimented with the following data:
\begin{enumerate}
\item All Mizar articles (MML version 1147), stripped of comments and concatenated together\footnote{\url{http://grid01.ciirc.cvut.cz/~mptp/nn_conj20/datasets/mmlall.txt2}}. This is 78M of uncompressed text.
\item Text version of the HTML export~\cite{Urban05} of the MML articles\footnote{\url{http://grid01.ciirc.cvut.cz/~mptp/nn_conj20/datasets/html2.tar.gz}}. This unpacks to 156MB. It additionally contains disambiguation features such as full types of variables, full names of theorems and the thesis is printed after every natural deduction step. This seems useful for neural conjecturing because the context is repeated more often.
\item Tokenized TPTP proofs\footnote{\url{http://grid01.ciirc.cvut.cz/~mptp/nn_conj20/datasets/prf2.tar.gz}} of 28271 Mizar theorems translated by the MPTP system~\cite{Urban06}. The proofs are produced by the E prover~\cite{Schulz13} equipped with recent ENIGMA guidance~\cite{DBLP:conf/cade/ChvalovskyJ0U19-short}. This unpacks to 658MB.
\item A subselection of the used Mizar premises from the 28271 proofs
  printed in prefix
  notation\footnote{\url{http://grid01.ciirc.cvut.cz/~mptp/nn_conj20/datasets/prf7.tar.gz}}. These
  files always start with the conjecture, and the premises are printed
  in the order in which E used them in its proof. This unpacks to 53MB.
\end{enumerate}
Below we show short examples of the four kinds of data, all for the theorem \texttt{ZMODUL01:103}:
\begin{footnotesize}
\begin{verbatim}
theorem
  for W being strict Submodule of V holds W /\ W = W
  proof
    let W be strict Submodule of V;
    the carrier of W = (the carrier of W) /\ (the carrier of W);
    hence thesis by Def15;
  end;

theorem :: ZMODUL01:103
for V being Z_Module
for W being strict Submodule of V holds W /\ W = W
proof
let V be Z_Module; ::_thesis: for W being strict Submodule of V holds W /\ W = W
let W be strict Submodule of V; ::_thesis: W /\ W = W
 the carrier of W = the carrier of W /\ the carrier of W ;
hence  W /\ W = W by Def15; ::_thesis: verum
end;

fof ( d15_zmodul01 , axiom , ! [ X1 ] : ( ( ( ( ( ( ( ( ( ( ~ ( v2_struct_0 ...
fof ( idempotence_k3_xboole_0 , axiom , ! [ X1 , X2 ] : k3_xboole_0 ( ...
fof ( t103_zmodul01 , conjecture , ! [ X1 ] : ( ( ( ( ( ( ( ( ( ( ~ ( ...
fof ( c_0_3 , plain , ! [ X118 , X119 , X120 , X121 ] : ( ( X121 ! = ...
cnf ( c_0_6 , plain , ( X1 = k7_zmodul01 ( X4 , X2 , X3 ) | v2_struct_0 ...

c! b0  c=> c& c~ cv2_struct_0 b0 c& cv13_algstr_0 b0 c& cv2_rlvect_1 b0 c& ...
c! b0  c=> c& c~ cv2_struct_0 b0 c& cv13_algstr_0 b0 c& cv2_rlvect_1 b0 c& ...
c! b0   c! b1  c= ck3_xboole_0 b0 b0 b0
\end{verbatim}
\end{footnotesize}

\section{Experiments}
The basic experiment for each dataset consists of training the
smallest (117M parameters) version of GPT-2 on a NVIDIA GeForce GTX
1080 GPU with 12GB RAM, producing random unconditioned samples during
the training. The produced samples and the most recent trained models
are available from our web
page\footnote{\url{http://grid01.ciirc.cvut.cz/~mptp/nn_conj20/samples/},
  \url{http://grid01.ciirc.cvut.cz/~mptp/nn_conj20/models/}}. The
published models can be used for conditional and unconditional
generation of Mizar-like texts, proofs and premise completion. The
samples contain megabytes of examples of what can be generated and how
the generated texts improve during the training. The training on the
third dataset was stopped early. The large number of redundant tokens
such as brackets and commas led us to produce the fourth dataset that
uses the punctuation-free prefix notation and much shorter summary of
the E proof (just the premises in their order). The training for
datasets 1, 2 and 4 has been running for several weeks, with the
performance still slowly improving. See Figure~\ref{chart} in Appendix~\ref{app:data}
for a sample training and loss on dataset 2.
There are many interesting conjectures generated during the
unconditioned sampling. The trained models can be directly used by Mizar users for autocompletion of their texts.
Some examples compared to real theorems are shown
below. More semantic evaluation on the textual datasets (1 and 2) could be done in various ways. We imagine that a proper Mizar environment will have to be guessed, some assumptions may be automatically added, etc.
\begin{footnotesize}
\begin{verbatim}
# real MML theorem
theorem :: YELLOW10:61
for S, T being non empty up-complete Poset
for X being Subset of S
for Y being Subset of T st X is property(S) & Y is property(S) holds
[:X,Y:] is property(S)

# generated similar "theorem"
theorem :: WAYBEL30:56
for R being complete Lattice
for X being Subset of [:R,R:] st X is upper & X is property(S) holds
[:X,X:] is property(S)

# real MML theorem
theorem :: SEQ_4:47
for X being non empty real-membered set
for Y being real-membered set st X c= Y & Y is bounded_below holds
lower_bound Y <= lower_bound X

# generated similar "theorem"
theorem Th33: :: SEQ_4:33
for X being real-membered set
for Y being real-membered set st Y is bounded_below & X c= Y holds
lower_bound Y <= lower_bound X

# generated "theorem" with "proof"
theorem Th23: :: STIRL2_1:23
for X, Y being finite set st not X is empty & X c= Y & 
card X = card Y holds X = Y
proof
let X, Y be finite set ; ::_thesis: ( not X is empty & X c= Y & 
card X = card Y implies X = Y )
assume that
A1: not X is empty and A2: X c= Y and A3: card X = card Y ; ::_thesis: X = Y
 card (Y \ X) = (card Y) - (card X) by A1, A3, CARD_2:44;
then A4: card (Y \ X) = ((card Y) - 1) - (card X) by CARD_1:30;
 X = Y \ X by A2, A3, Th22;
hence  X = Y by A4, XBOOLE_0:def_10; ::_thesis: verum
end;
\end{verbatim}
\end{footnotesize}

\subsection{Conditioned completion}
We have also done two initial experiments with proof and formula
completion. The data and results for them are available from our web
page\footnote{\url{http://grid01.ciirc.cvut.cz/~mptp/nn_conj20/samples/premises/},
  \url{http://grid01.ciirc.cvut.cz/~mptp/nn_conj20/samples/html2/}}.
In the first experiment, we use a model trained on Dataset 4
(premises), and ask the model to auto-complete 369 theorems from the
CARD series of Mizar. For each conjecture we produce 10 premise
selections using beam search, and we use different temperatures and
beam search parameters. An interesting phenomenon is that with low
temperatures, practically all conjectured premises are known Mizar
theorems. I.e., the task reduces to standard premise selection. With
higher temperatures, GPT-2 starts producing premises (lemmas) that are
not among the existing Mizar theorems, but are still well-typed. Even
higher temperatures lead to non-well-typed or even unparsable lemmas.
The next section provides a more involved ATP evaluation done on a larger dataset.

The second experiment was done over Dataset 2 and a set of 462 partial formulas from the CARD articles.
The model trained on Dataset 2 is then (again using beam search) asked to auto-complete these formulas.
Mizar users can also play with such autocompletion via a web server\footnote{\url{http://grid01.ciirc.cvut.cz:8000/}} using this model.
For example,
\begin{footnotesize}
\begin{verbatim}
for M, N being Cardinal holds
\end{verbatim}
\end{footnotesize}
results in the following (selected) auto-completions (more are available online\footnote{\url{http://grid01.ciirc.cvut.cz/~mptp/nn_conj20/samples/html2/00cardmizout1_t1}}):
\begin{footnotesize}
\begin{verbatim}
( M = N iff M,N are_equipotent )
( M = N iff not 0 in M ) by ORDINAL3:8;
M *` N = N *` M
( M in N iff not M c= N )
the_rank_of M = the_rank_of N by Th77;
nextcard (Sum M) = M *` N
\end{verbatim}
\end{footnotesize}

\subsection{Initial ATP Evaluation}
The first larger ATP (semantic) evaluation uses the fourth dataset following the setting  introduced for such evaluations in~\cite{GauthierKU16-short}. 
After training GPT-2 on the 28271 ENIGMA proofs, we produce (using beam search) 12 GPT-2 premise predictions for a set of 31792 theorems of which 6639 are not among the training ones.
This yields 381432 predictions,\footnote{\url{http://grid01.ciirc.cvut.cz/~mptp/nn_conj20/results/preds3.tar.gz}}
deduplicated to 193320 unique predictions.
The predictions are converted back to TPTP from the polish
notation, creating ATP problems. We distinguish between the premises
that already exist as Mizar theorems and definitions, and the new
formulas (conjectures) introduced by GPT-2. 108564\footnote{\url{http://grid01.ciirc.cvut.cz/~mptp/nn_conj20/results/preds5.tar.gz}} of the created
problems contain no new conjectures, i.e., GPT-2 works there as a
standard premise selector similar to~\cite{piotrowski2020stateful}.

Most (86899) of these ATP
problems\footnote{\url{http://grid01.ciirc.cvut.cz/~mptp/nn_conj20/results/preds6.tar.gz}}
can be quickly shown to be countersatisfiable by E prover.\footnote{We
  used E with 6 s time limit and its auto-schedule mode for this
  initial check.} This shows the first difference between syntactic
loss as used by the ML/NLP community and semantic usefulness. GPT-2's
loss is geared towards mimicking the length of the original texts with
a small number of syntactic mistakes. In premise selection, the
underlying task is to generate premises that have sufficient logical
power. Overshooting is better than making a mistake and observing the
usual length of the text.
11866 of the problems can be proved in 6 s, resulting in
proofs of 8105 theorems. This is not yet an interesting number,
because GPT-2 does not observe the
\emph{chronological order} of premises. E.g., 4350 of the proofs use
only a single premise -- typically GPT-2 suggested the proved theorem itself as a premise.
Still, some predictions are chronologically correct and lead to
correct new proofs. E.g. for theorem
XXREAL_1:48,\footnote{\url{http://grid01.ciirc.cvut.cz/~mptp/7.13.01_4.181.1147/html/xxreal_1.html\#T48}}
which is not in the training set, the fifth GPT-2 sample proposed 7 premises\footnote{\url{http://grid01.ciirc.cvut.cz/~mptp/nn_conj20/results/t48_xxreal_1___5}} of which 5 were used in a quickly found new E proof\footnote{\url{http://grid01.ciirc.cvut.cz/~mptp/nn_conj20/results/t48_xxreal_1___5.out}}
(see Appendix~\ref{app:data} for details).

Next we
evaluate\footnote{\url{http://grid01.ciirc.cvut.cz/~mptp/nn_conj20/results/preddatagpt1.out.tar.gz}}
the 44524
problems\footnote{\url{http://grid01.ciirc.cvut.cz/~mptp/nn_conj20/results/preddatagpt1.tar.gz}}
that do use at least one newly proposed premise. We have not strictly
enforced the chronology, but remove the theorem itself from axioms if
proposed. 34675 of the problems are then found countersatisfiable by E
in 1 s and for 1515 a proof is found. The conjectures may be
interesting, even though hard to prove automatically: E.g. for
GROUPP_1:T10\footnote{\url{http://grid01.ciirc.cvut.cz/~mptp/7.13.01_4.181.1147/html/groupp_1.html\#T10}}
a valid, though not quite trivial strengthening from finite to general
groups is proposed, see Appendix~\ref{app:data} for details.

In total, GPT-2 proposed in this experiment 52515
new syntactically correct
formulas\footnote{\url{http://grid01.ciirc.cvut.cz/~mptp/nn_conj20/results/out4.tar.gz}}
that deduplicate to 33100.
Some are clearly false, yet quite natural to ask: e.g. for dozens of
theorems like
SINCOS10:17\footnote{\url{http://grid01.ciirc.cvut.cz/~mptp/7.13.01_4.181.1147/html/sincos10.html\#T17}}
-- ``\texttt{sec is increasing on $[0,\pi / 2)$}'' -- GPT-2 makes the
conjecture that every differentiable function is
increasing.\footnote{\url{http://grid01.ciirc.cvut.cz/~mptp/nn_conj20/results/t17_sincos10___1}}
In this particular case we can likely disprove the conjecture since
there are counterexamples in the MML. Similarly, in
FUNCTOR1:9\footnote{\url{http://grid01.ciirc.cvut.cz/~mptp/7.13.01_4.181.1147/html/functor1.html\#T9}},
to prove that the composition of full functors is full, GPT-2 proposes
to reduce fullness to faithfulness, likely because a previous
theorem\footnote{\url{http://grid01.ciirc.cvut.cz/~mptp/7.13.01_4.181.1147/html/functor1.html\#T7}}
says that faithfulness is preserved under composition. See Appendix~\ref{app:data} for details.

Finally we use standard premise selection (although we could
recurse and use GPT-2) and E with the ENIGMA guidance to try to prove the 52515 new
formulas.\footnote{\url{http://grid01.ciirc.cvut.cz/~mptp/nn_conj20/results/preddata128.tar.gz}}
This yields 9000-10000
proofs,\footnote{\url{http://grid01.ciirc.cvut.cz/~mptp/nn_conj20/results/preddata128.out.tar.gz}}
depending on how we run premise selection and E. While some proofs are long, it seems that
we are not yet capable of proving the more interesting conjectures and we still need more ATP strengths. E.g., the longest ATP proof shows that \texttt{-infty is non empty}, where \texttt{-infty} is defined as \texttt{[0,REAL]}. A slightly more useful conjecture which is also hard to prove\footnote{\url{http://grid01.ciirc.cvut.cz/~mptp/nn_conj20/results/t20_borsuk_3___7__1}} is the strengthening of the symmetry of the \texttt{are_homeomorphic} predicate\footnote{\url{http://grid01.ciirc.cvut.cz/~mptp/7.13.01_4.181.1147/html/borsuk_3.html\#R2}} from non-empty to arbitrary spaces.

\bibliographystyle{plain}
\bibliography{vectconj,ate20}

\appendix
\section{Additional Data From the Experiments}
\label{app:data}
\begin{figure}[htbp!]
\begin{center}
\includegraphics[scale=0.4]{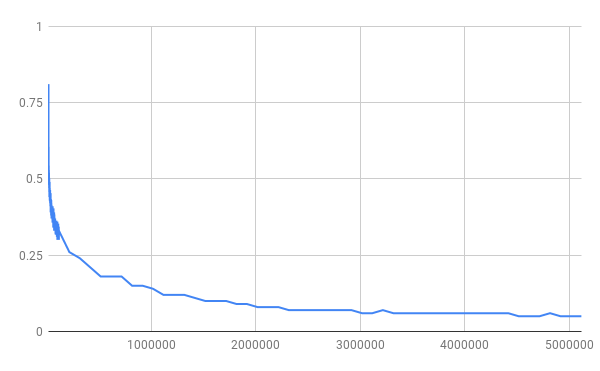}
\end{center}
\caption{\label{chart}Dataset 2 training and loss.}
\end{figure}

\subsection{\texttt{XXREAL 1:48} and its GPT-2 predictions}

\begin{lstlisting}[language=Mizar,basicstyle=\small]
theorem Th48: :: XXREAL_1:48
for p, r, s, q being ext-real number st p < r & s <= q holds
[.r,s.[ c= ].p,q.[
\end{lstlisting}
Following are the Mizar premises in the order proposed by GPT-2. The fifth and sixth were not needed for the ATP proof.
\begin{lstlisting}[language=Mizar,basicstyle=\small]

theorem Th3: :: XXREAL_1:3
for t, r, s being ext-real number holds
t in [.r,s.[ iff  r <= t & t < s  

let X be ext-real-membered set ; let Y be set ;
pred X c= Y means :Def8: :: MEMBERED:def 8
for e being ext-real number st e in X holds e in Y;

let r, s be ext-real number ;
cluster [.r,s.[ -> ext-real-membered ;

theorem Th2: :: XXREAL_0:2
for a, b, c being ext-real number st a <= b & b <= c holds a <= c

let X be ext-real-membered set ;
cluster -> ext-real for Element of X;

theorem :: SUBSET:1
for a, b being set st a in b holds a is Element of b;

theorem Th4: :: XXREAL_1:4
for t, r, s being ext-real number holds
t in ].r,s.[ iff  r < t & t < s 

\end{lstlisting}

\subsection{\texttt{GROUPP_1:10} and its generalization conjectured by GPT-2}

\begin{lstlisting}[language=Mizar,basicstyle=\small]
theorem Th10: :: GROUPP_1:10
for G being finite Group for N being normal Subgroup of G st
N is Subgroup of center G & G ./. N is cyclic holds
G is commutative
\end{lstlisting}
The generalization that avoids finiteness:

\begin{lstlisting}[language=Mizar,basicstyle=\small]
for G being Group for N being normal Subgroup of G st
N is Subgroup of center G & G ./. N is cyclic holds
G is commutative
\end{lstlisting}
We don't have an ATP proof of the generalization yet. We thank
algebraists Michael Kinyon and David Stanovsk\'y for confirming that
this generalization is provable. Based on this example Stanovsk\'y
commented that related Mizar theorems can be similarly
generalized.

\vspace{2mm}
\subsection{\texttt{SINCOS10:17} and a false conjecture by GPT-2}

\begin{lstlisting}[language=Mizar,basicstyle=\small]
theorem Th17: :: SINCOS10:17
sec | [.0,(PI / 2).[ is increasing
\end{lstlisting}
GPT-2 generated the following conjecture, which is false. Along with another GPT-2 conjecture about the differentiability of \texttt{sec} on the interval,  this results in an ATP proof of \texttt{SINCOS10:17}.

\begin{lstlisting}[language=Mizar,basicstyle=\small]
for X being set for f being Function of REAL, REAL holds
f is differentiable_on X implies f | X is increasing
\end{lstlisting}

\subsection{\texttt{FUNCTOR1:9} and a GPT-2 conjecture reducing it to \texttt{FUNCTOR1:7}}
\begin{lstlisting}[language=Mizar,basicstyle=\small]
theorem Th9: :: FUNCTOR1:9
for C1 being non empty AltGraph
for C2, C3 being non empty reflexive AltGraph
for F being feasible FunctorStr over C1,C2
for G being FunctorStr over C2,C3
st F is full & G is full holds G * F is full

for C1, C2 being AltGraph for F being FunctorStr over C1,C2
holds F is full iff F is faithful & F is feasible

theorem Th7: :: FUNCTOR1:7

for C1 being non empty AltGraph
for C2, C3 being non empty reflexive AltGraph
for F being feasible FunctorStr over C1,C2
for G being FunctorStr over C2,C3
st F is faithful & G is faithful holds G * F is faithful

\end{lstlisting}

\end{document}